\documentclass[letterpaper, 10 pt, conference]{ieeeconf}  % Comment this line out if you need a4paper

\IEEEoverridecommandlockouts                              % This command is only needed if 
\usepackage{graphics} % for pdf, bitmapped graphics files
\usepackage{epsfig} % for postscript graphics files
\usepackage{mathptmx} % assumes new font selection scheme installed
\usepackage{times} % assumes new font selection scheme installed
\usepackage{amsmath} % assumes amsmath package installed
\usepackage{amssymb}  % assumes amsmath package installed
\usepackage{graphicx}
\usepackage{verbatim}
\usepackage[table]{xcolor}
\usepackage{multirow}

\title{\LARGE \bf
Learning a Vision-Based Footstep Planner for Hierarchical Walking Control}

\author{Minku Kim, Brian Acosta, Pratik Chaudhari and Michael Posa% <-this % stops a space
\thanks{The authors are with the General Robotics, Automation, Sensing and Perception (GRASP) Laboratory, University of Pennsylvania, Philadelphia, PA, 19104, USA {\tt\small \{minkukim, bjacosta, pratikac, posa\}@seas.upenn.edu}}%
}

\begin{document}

\maketitle
\thispagestyle{empty}
\pagestyle{empty}

%%%%%%%%%%%%%%%%%%%%%%%%%%%%%%%%%%%%%%%%%%%%%%%%%%%%%%%%%%%%%%%%%%%%%%%%%%%%%%%%
\begin{abstract}
Bipedal robots demonstrate potential in navigating challenging terrains through dynamic ground contact. However, current frameworks often depend solely on proprioception or use manually designed visual pipelines, which are fragile in real-world settings and complicate real-time footstep planning in unstructured environments. To address this problem, we present a vision-based hierarchical control framework that integrates a reinforcement learning high-level footstep planner, which generates footstep commands based on a local elevation map, with a low-level Operational Space Controller that tracks the generated trajectories. We utilize the Angular Momentum Linear Inverted Pendulum model to construct a low-dimensional state representation to capture an informative encoding of the dynamics while reducing complexity. We evaluate our method across different terrain conditions using the underactuated bipedal robot Cassie and investigate the capabilities and challenges of our approach through simulation and hardware experiments.
\end{abstract}
%\vspace{2mm} \Note{Using the Note command will make the enclosed text red}
%%%%%%%%%%%%%%%%%%%%%%%%%%%%%%%%%%%%%%%%%%%%%%%%%%%%%%%%%%%%%%%%%%%%%%%%%%%%%%%%
\section{Introduction}
Bipedal robots hold immense potential for traversing unstructured terrains, making them invaluable for applications such as search and rescue and disaster response \cite{yoshiike2017development}. Humans demonstrate remarkable adaptive locomotion through a combination of proprioceptive feedback and visual perception. When walking, we simultaneously evaluate our surroundings for safe and stable footholds while planning our next step. This seamless integration of sensory information and control is essential for effective navigation in outdoor environments.
%When walking on challenging terrain, such as hiking trails,

To replicate this behavior, vision-based locomotion controllers generally follow a modular framework composed of perception, planning, and control. The perception module processes visual data to build a spatial map of the local environment, which then informs a high-level planner to determine foot placements or motion trajectories. Finally, a low-level controller translates these plans into actuator commands. Although this pipeline has demonstrated success in structured settings, it often relies on handcrafted visual features and model-based strategies that are sensitive to noise and typically limited to simplified terrain assumptions such as piecewise-planar surfaces \cite{deits2014footstep, fallon2015continuous, kim2020vision, 11077715}.

An important component in achieving tractable planning for bipedal locomotion is the use of reduced-order models (ROMs), which are simplified representations that capture the key dynamics of complex robotic systems. These models offer interpretable and computationally efficient formulations, making them useful for real-time motion planning and control. Although ROMs have traditionally been employed in model-based control frameworks \cite{xiong2018coupling, gong2022zero}, recent research has demonstrated their usefulness within reinforcement learning frameworks as well \cite{green2021learning, castillo2023template, chen2024reinforcement}.

Recent developments in reinforcement learning (RL) have significantly advanced the integration of visual perception and locomotion control by enabling both end-to-end control policies \cite{miki2022learning, duan2024learning} and hybrid control architectures \cite{yu2021visual, gangapurwala2022rloc} that operate directly on raw or minimally processed visual inputs. These approaches reduce the reliance on manually engineered features, thereby increasing flexibility and generalization. However, this introduces challenges in transferring policies from simulation to real-world hardware.

\begin{figure}[t]
    \centering
    \includegraphics[width=\linewidth]{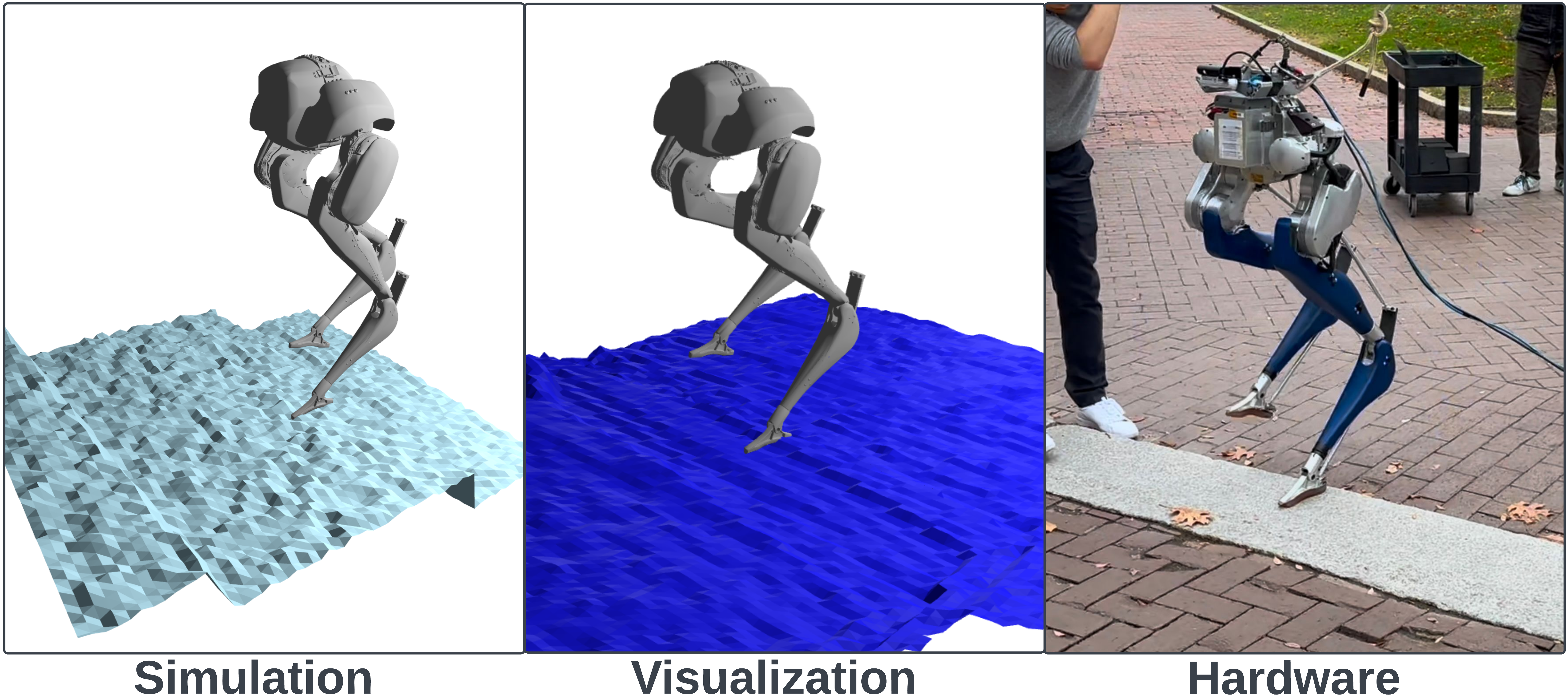}
    \vspace{-7mm}
    \caption{The system incorporates local terrain information through an elevation map, enabling vision-based footstep planning via a reinforcement learning (RL) policy trained in simulation. The approach is validated on hardware. Left: Cassie training in simulation. Center: Visualization of the real-world elevation map. Right: Hardware evaluation.}
    \vspace{-5mm}
    \label{fig:cassie}
\end{figure}

Inspired by recent works, we propose a vision-based hierarchical control framework that integrates an RL-based footstep planner with a low-level operational space controller (OSC). The RL policy utilizes visual and low-dimensional state representation inputs to generate 3D footstep placements in real-time, while the low-level OSC tracks the spline trajectories derived from these footsteps.

The main contributions of this paper are:
\vspace{-2mm}
\begin{itemize}
    \item A vision-based hierarchical controller that uses a single depth camera and a reduced-order model to enable efficient and interpretable 3D footstep planning via reinforcement learning.

    \item Hardware validation of the full pipeline on both structured and unstructured terrains, with benchmarking against a model predictive control (MPC) baseline in simulation.

    \item An analysis of the ALIP model as a policy input and hierarchical control structure, showing limitations on complex terrain and impact on sim-to-real transfer.
\end{itemize}

% We validate the full pipeline in real-world hardware experiments across both structured and unstructured terrains, and benchmark it against a Model Predictive Control (MPC) baseline in simulation. 

% We analyze the impact of using the ALIP model as policy input and hierarchical control structure, showing the limitations in handling complex terrains such as stairs for the ALIP model and the hierarchical structure complicates sim-to-real transfer.

% We analyze the impact of using the ALIP model as policy input and hierarchical control structure, showing limitations in handling complex terrains, and showing that the hierarchical architecture further complicates sim-to-real transfer.

% limitations introduced by the use of a reduced-order ALIP model and a hierarchical control architecture. While these choices simplify learning and improve interpretability, the ALIP model’s limited representational capacity hinders performance on complex tasks such as stair climbing. Moreover, the hierarchical structure complicates sim-to-real transfer by introducing inter-layer dependencies and coordination challenges, which reduce policy generalization in real-world deployment.

% Analysis of the impact of using a reduced-order model as policy input and decision structure, showing limitations in handling complex terrain such as stairs due to restricted representational capacity.

\section{Related Work}

\subsection{Optimal Control for Bipedal Locomotion}
Model-based control for bipedal locomotion is often structured as an optimal control problem, where the multi-body dynamics are embedded as constraints in the control formulation \cite{wensing2023optimization}. However, due to the high dimensionality of full-order dynamics in bipedal robots, this formulation becomes computationally impractical for real-time optimization. Consequently, simplified reduced-order models, such as the Linear Inverted Pendulum (LIP) model \cite{kajita20013d} are used for online control along with its variations, including SLIP \cite{rummel2010stable}, ALIP \cite{gong2022zero}, and H-LIP \cite{xiong20223}. Controllers that use reduced-order dynamics with Model Predictive Control (MPC) for footstep planning have been successfully deployed in real-world environments \cite{gibson2022terrain}. However, there still remains limited investigation into robust vision-based control strategies for bipedal robots in uneven terrain, where previous methods have lacked robustness against noisy visual inputs such as poor lighting, making the pipeline unreliable \cite{deits2014footstep, fallon2015continuous, acosta2023bipedal}.

\subsection{Reinforcement Learning for Bipedal Locomotion}
In response to the limitations of traditional control methods, recent research has actively explored deep reinforcement learning (DRL) as an alternative. Leveraging advancements in computation and physics simulations, DRL has enabled robust walking controllers capable of navigating diverse terrains in simulation and real-world environments \cite{xie2018feedback, siekmann2021blind} and learning unified frameworks that can perform various dynamic gaits such as walking, running, and jumping while maintaining robustness against perturbations \cite{li2024reinforcement}. Recent work has also shown that hierarchical control architectures that combine model-based methods with DRL can enhance generalizability, interpretability, and sample efficiency by decomposing the locomotion problem into separate layers of objectives \cite{castillo2023template, yu2021visual, gangapurwala2022rloc, bao2024deep}. Learning approaches demonstrate better performance in visual-locomotion integration compared to optimal control methods for both quadrupeds and bipeds, primarily due to their capacity to establish direct mappings between visual inputs and motor commands or footstep planning \cite{miki2022learning, duan2024learning, miki2022learning, agarwal2023legged}.

\subsection{Sim-To-Real Transfer}
Despite the promising capabilities of DRL in developing locomotion controllers, several challenges emerge when deploying these controllers on physical robots. The primary limitation is due to modeling errors between the simulated environment and the real-world environment, making direct transfer of policies from simulation to hardware difficult. Domain randomization \cite{tobin2017domain} has emerged as a prevalent approach to bridge this reality gap. Instead of carefully tuning model parameters to match real-world conditions, domain randomization involves extensively randomizing the simulated environment. By exposing the policy to a range of model distributions, the policy learns the distribution shift between the real-world and simulated environment \cite{siekmann2021blind, li2021reinforcement}.
\begin{figure*}[t]
\centering
% \vspace{-3mm}
\includegraphics[width=\linewidth]{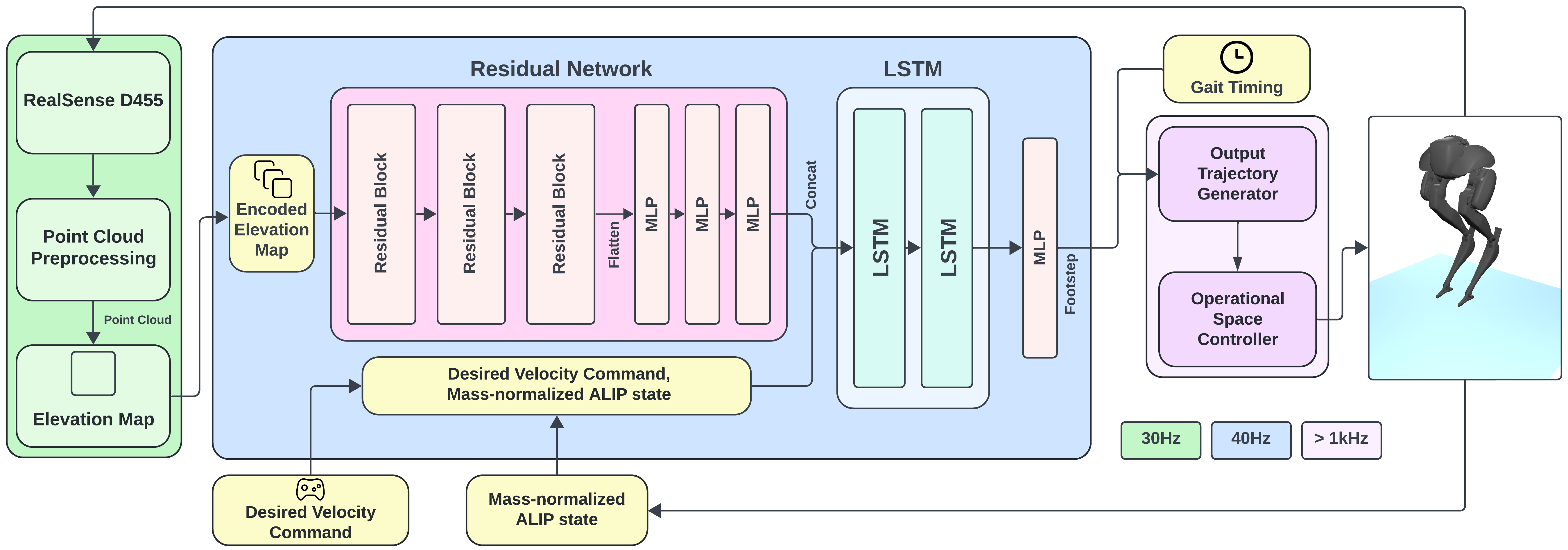}
\caption{Overview of the system diagram. The perception module (green) generates elevation maps at 30Hz from a RealSense D455. The high-level footstep policy (blue) outputs footstep actions at 40Hz. These actions are sent to the low-level controller (purple) for joint control.}
\label{fig:overview}
\vspace{-5mm}
\end{figure*}

\section{Background}
\subsection{Angular Momentum Linear Inverted Pendulum model} The classical Linear Inverted Pendulum (LIP) model \cite{kajita20013d} simplifies legged robots to a point mass at the center of mass (CoM) supported by a massless leg.
The Angular Momentum Linear Inverted Pendulum (ALIP) model \cite{gong2022zero} is a reparameterization of the traditional LIP model, where the linear velocity of the CoM is replaced by the angular momentum about the contact point as the velocity variable. In the idealized case, the ALIP and LIP are mathematically equivalent. However, for real robotic systems with distributed mass and articulated limbs, the ALIP model is less sensitive to model error introduced by the movement of these limbs. In this work, we use a mass-normalized ALIP model, where the angular momentum is divided by the robot's mass to give the state components similar magnitudes:
\[
\dot{x} = 
\begin{bmatrix} 
    \dot{x}_{\text{com}} \\ \dot{y}_{\text{com}} \\ \dot{l}_x \\ \dot{l}_y 
\end{bmatrix} = 
\begin{bmatrix} 
    0 & 0 & 0 & \frac{1}{H} \\ 
    0 & 0 & -\frac{1}{H} & 0 \\ 
    0 & -g & 0 & 0 \\ 
    g & 0 & 0 & 0 
\end{bmatrix}
\begin{bmatrix}
    x_{\text{com}} \\ y_{\text{com}} \\ l_x \\ l_y
\end{bmatrix}
%+ \begin{bmatrix}
%    0 \\ 0 \\ 0 \\ 1
%\end{bmatrix}u
\]
where $x_{CoM}$ and $y_{CoM}$ represent the horizontal positions of the CoM in the x and y directions, and $l_x$ and $l_y$ are the mass-normalized horizontal components of the angular momentum about the contact point in the x and y directions, and $H$ denotes the height of the CoM.% For a detailed derivation of the ALIP dynamics, we direct readers to \cite{gibson2022terrain}.

\subsection{Reinforcement Learning}
In a standard reinforcement learning (RL) task, an agent engages in a sequential decision-making process, interacting with an environment by observing the current state, selecting actions based on a policy, and receiving rewards that guide future decisions. The problem is modeled using a Markov Decision Process (MDP) defined by a tuple of $(S, A, P, R, \gamma)$, where $S$ represents a set of states $s$, $A$ denotes a set of actions $a$. $P(s'|s,a)$ describes the probability of transitioning from the current state $s$ to the next state $s'$ when an action $a$ is taken. $R(s,a)$ gives an immediate scalar reward for each transition made from a state and action pair.
The goal of RL is to learn an optimal policy $\pi^*_\theta$ that maximizes the expected value of the cumulative rewards and is formalized as:
\begin{equation}
\max_{\theta} J(\theta;s_0) = \mathbb{E} \left[ \sum_{t=0}^{\infty} \gamma^t R(s_t, a_t) | s_0 \right]
\end{equation}
%$P(s'|s,a)$ is the state transition probability function, $R(s,a)$ is the reward function, and $\gamma$ is the discount factor that balances the short-term and long-term rewards.
% The actor-critic method involves two components: the actor and the critic. The actor represents a parameterized policy $\pi(a|s;\theta)$, where $\theta$ denotes the parameters of the policy, and updates by learning from the feedback provided by the critic. The critic evaluates the actions taken from the actor by estimating the value function.

\section{Learning and Control Architecture}
% \textbf{Figure with the System Diagram}
\begin{figure}[t]
    \centering
    \includegraphics[width=\linewidth]{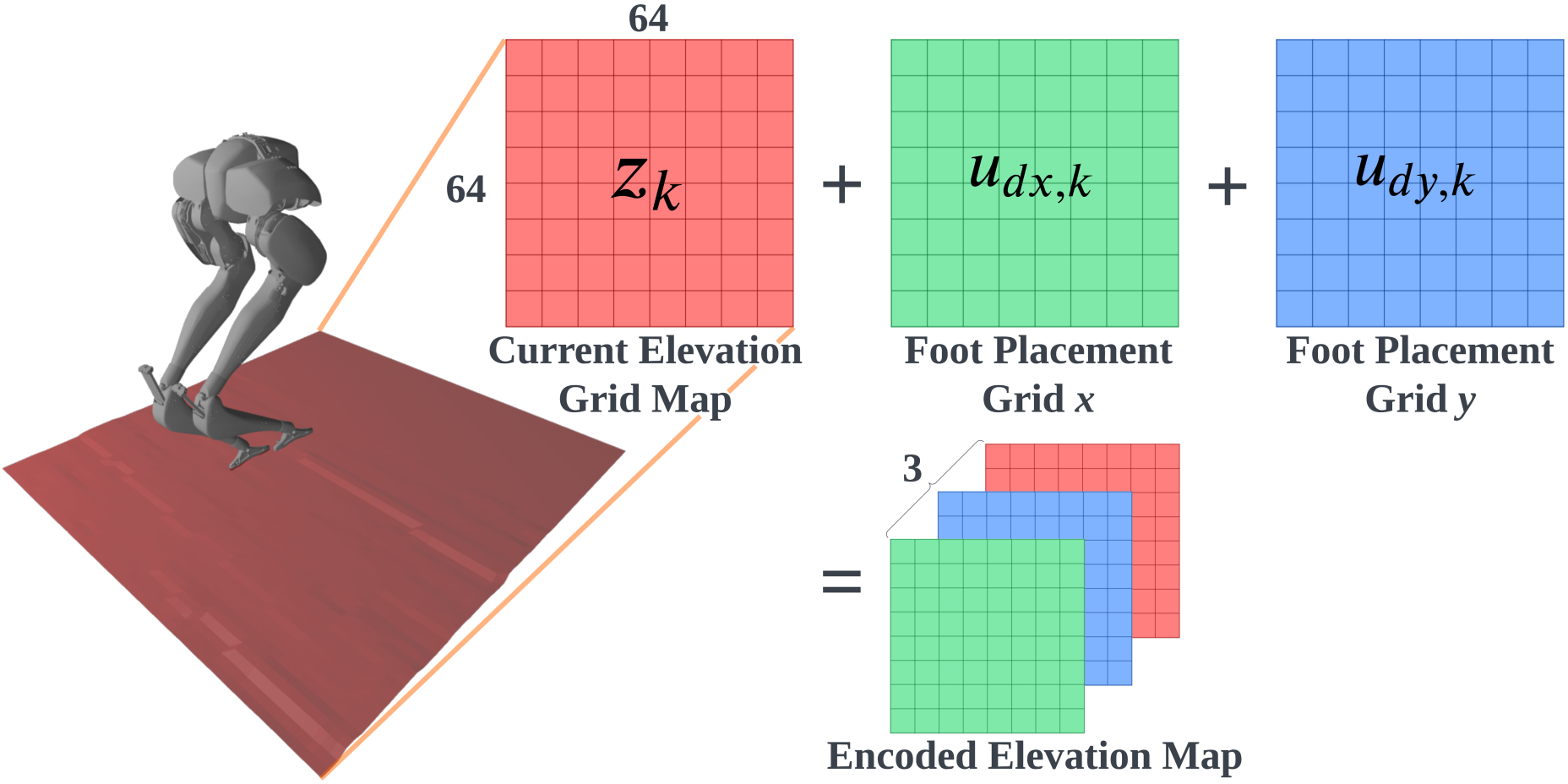}
    \vspace{-7mm}
    \caption{The elevation map is cropped to a 64$\times$64 grid and concatenated with XY footstep location grids. The central position of these grids corresponds to the desired footstep placement, which is calculated using an ALIP trajectory based on the velocity command and a constant stance width of 20 cm.}
    \vspace{-5mm}
    \label{fig:encode}
\end{figure}
\subsection{Perception Module}
We use an Intel RealSense D455 to provide point-cloud updates to a robot-centric elevation mapping framework \cite{fankhauser2018probabilistic}. 
The D455 is mounted to the pelvis and pointed downward in front of the robot. 
We pre-process the point clouds to mask out the robot's legs before inputting the point clouds to the elevation map.  
The map is updated at 30 Hz based on the most recent point-cloud measurements. 
The robot state is estimated by a contact-aided invariant EKF \cite{hartleyINEKF}, which uses simulated sensor measurements (encoder and IMU values) during training, and real sensor measurements when deployed on hardware. 
This state estimate is also used to compute the ALIP state observation.
The map is cropped into a 64$\times$64 grid with a resolution of 0.025 meters per cell, which provides sufficient terrain detail and aligns with the range of the camera.
Unknown heights are filled via nearest-value interpolation, and a median filter smooths out the values. 
The elevation map is tiled with a grid of potential footstep locations, centered with the desired footstep position ($u_{dx}$, $u_{dy}$) computed from a periodic ALIP trajectory to provide structured spatial information (Fig. \ref{fig:encode}).

\subsection{Reinforcement Learning for Footstep Planning}
The RL footstep planner uses the elevation map, ALIP states, and user-provided velocity commands to determine the footstep location local to the stance frame. The RL footstep planner is trained in simulation and then transferred to the physical robot. The policy is first pre-trained with ALIP trajectories from a blind Linear Quadratic Regulator (LQR) controller before being fine-tuned with model-free RL. The training process randomizes the terrain, velocity command, and model parameters using a curriculum learning approach to facilitate stable training and sim-to-real transfer.

\subsection{Task-Space Objectives for Whole-Body QP}
To realize the output of the learned footstep planner on the robot, we use an inverse-dynamics operational-space control QP~\cite{wensing2013generation}. The gains can be seen in Table \ref{tab:osc_sim_gains}, where the weight and gain matrices are diagonal and are represented as vectors. The swing foot trajectory is represented as a single-segment polynomial spline that ends at the target footstep location. 
This spline is replanned for every new footstep target with a quadratic program (QP) that ensures continuity of the desired position, velocity, and acceleration, as well as minimizing swing foot acceleration and distance of the spline's midpoint from a target midpoint, which ensures ground clearance \cite{khadiv2020walking}.
The center of mass height is controlled to a virtual plane that defines the ALIP model for the current and upcoming stance foot.
The swing foot angle is also controlled to be parallel to this plane. 
To fully control Cassie's remaining degrees of freedom, we control the pelvis pitch and roll to zero, the swing leg yaw angle to zero, and the pelvis yaw rate to match a turning rate commanded by an operator.
The commanded pelvis yaw rate is zero during training.

\begin{table}[t]
\scriptsize
\centering
\caption{Feedback gains for the OSC controller}
\vspace{-3mm}
\label{tab:osc_sim_gains}
\begin{tabular}{|l|c|c|c|}
\hline
\textbf{OSC Objective} & $\mathbf{W}$ & $\mathbf{K_p}$ & $\mathbf{K_d}$ \\
\hline \hline
Toe joint angle       & 1         & 1500         & 10     \\
\hline
Hip yaw angle         & 2         & 100          & 4      \\
\hline
Pelvis [x, y]         & [2, 4]    & [200, 200]   & [10, 10] \\
\hline
Pelvis heading [yaw]  & 0.02      & 0            & 10     \\
\hline
CoM [z]               & 10        & 80           & 5      \\
\hline
Foot [x, y, z]        & [4, 4, 2] & [400, 400, 400] & [20, 20, 25] \\
\hline
\end{tabular}
\vspace{-5mm}
\end{table}
\section{Learning a Vision-based Footstep Planner}
\subsection{Policy Design}
The policy $\pi_\theta$ takes as inputs the desired velocity $v_{des}$ provided by the operator, the mass-normalized ALIP state $s_{ALIP}$, and an encoded elevation map $M_{elevation}$ and outputs the foot placement coordinates in the local stance frame $[p_x, p_y, p_z]$.  The use of task space actions improves sample efficiency and enables broader exploration to discover solutions compared to joint space actions, as shown in \cite{duan2021learning}.

The policy architecture consists of two main components (Fig. \ref{fig:overview}): a residual network and a Long Short-Term Memory (LSTM) network. The residual network processes the encoded elevation map $M_{elevation}$ to produce a latent representation, which is combined with the current observation vector $[v_{des}, s_{ALIP}]$. This serves as input to the LSTM network, which generates the actions. Prior to sending the actions out, the actions are concatenated with gait timing information to ensure temporal consistency with the gait cycle. The actor and critic networks do not share any layers.

\begin{table}[t]
\scriptsize
\centering
\caption{Reward and Penalty Terms}
% \vspace{-3mm}
\label{table:terms}
\begin{tabular}{|c||l|c|}
\hline
\textbf{Terms}                    & \textbf{Name} & \textbf{Value}           \\ \hline
\multirow{6}{*}{Reward}  & Forward Velocity $(r_{v_x})$ & $0.5 \times e^{-2\|v_{x,des} - v_x\|}$\\                                 \cline{2-3} 
                         & Lateral Velocity $(r_{v_y})$ & $0.25 \times e^{-2\|v_{y,des} - v_y\|}$\\
                         \cline{2-3} 
                         & Height $(r_z)$ & $0.3125 \times e^{-4\|a_{z,GT} - a_z\|}$\\
                         \cline{2-3} 
                         & Pelvis Stability $(r_{\phi})$ & $0.1875 \times e^{-2\| \omega_z \|}$\\
                         \cline{2-3} 
                         & Action Smoothness $(r_{a_t})$ & $0.125 \times e^{-3\|a_{t} - a_{t-1}\|}$\\
                         \cline{2-3}
                         & Action Regulation $(r_{reg})$ & $0.125 \times e^{-2\|a_{x,y,des} - a_{x,y}\|}$\\
                         \hline

\multirow{3}{*}{Penalty} & Tracking Penalty  $(p_{track})$        & $e^{3.5(err-0.05)}-1$ \\
                         \cline{2-3}
                         & Torque Penalty   $(p_{\tau})$       & $-0.000007 \times \sum_{i} \tau_i^2$ \\
                         \cline{2-3}
                         & Edge Penalty   $(p_{edge})$      & $1.5 \times \mathbf{1}_{\{step \hspace{0.3em} on \hspace{0.3em} edge\}}$ \\
                         \cline{2-3}
                         & Collision Penalty  $(p_{collision})$       & $1.5 \times \mathbf{1}_{\{collision\hspace{0.3em} with\hspace{0.3em} terrain\}}$   \\
                        \hline
\end{tabular}
\vspace{-5mm}
\end{table}

\subsection{Reward Function}
We denote $v$ as the linear velocity, $a$ as the footstep action, $\omega$ as the angular velocity, and $\tau$ as the torque. The reward function is the sum of the reward terms and penalty terms shown in Table \ref{table:terms}. The total reward is constrained to be non-negative; If the cumulative penalty exceeds the cumulative reward, the total reward is set to zero. This design choice helps prevent the destabilizing effects of large penalties and promotes more stable policy learning.
\begin{align}
r = \min\Big( 
    &\mathbf{w}_r^\top 
    \begin{bmatrix}
        r_{v_x} \\
        r_{v_y} \\
        r_{z} \\
        r_{\phi} \\
        r_{a_t} \\
        r_{reg}
    \end{bmatrix}
    -
    \mathbf{w}_p^\top 
    \begin{bmatrix}
        p_{track} \\
        p_{\tau} \\
        p_{edge} \\
        p_{collision}
    \end{bmatrix},
    \; 0 \Big)
\end{align}
% \begin{align}
%    r &= \text{min}( w_{r}^T \begin{bmatrix} r_{v_x}, r_{v_y}, r_{\phi}, r_{z}, r_{a_{t}}, r_{reg} \end{bmatrix}^T \\
%    &- w_{p}^T \begin{bmatrix} p_{track}, p_{\tau}, p_{edge}, p_{collision} \end{bmatrix}^T, 0) \notag
% \end{align}
The reward terms $r_{v_x}$ and $r_{v_y}$ encourage the policy to track the desired velocities, while $r_{\phi}$ promotes pelvis stability by minimizing yaw angular velocity. The $r_z$ term aligns footstep heights with terrain elevation, and $r_{a_t}$ encourages action continuity within the same stance. Finally, $r_{reg}$ regulates actions to remain close to optimal footsteps determined by the LQR equation.
The penalty term $p_{track}$ penalizes deviations in swing foot trajectory tracking, $p_{\tau}$ penalizes torque usage to discourage the policy from generating excessively large torque commands, $p_{edge}$ discourages stepping near edges detected by a Sobel filter \cite{kanopoulos1988design}, and $p_{collision}$ penalizes front foot collisions with the terrain.

Excluding edge penalties led to foot placements near stair edges, resulting in frequent slips and elevation map artifacts caused by the drift correction. Without the collision penalties, Cassie developed a toe-probing strategy before stepping up. While this strategy is viable in simulation, this behavior results in hardware failures.

%Excluding edge and collision penalties during training introduced suboptimal behaviors. The absence of edge penalties resulted in foot placements near stair edges, causing slips and elevation map artifacts caused by incorrect height interpretations during drift correction. Without collision penalties, Cassie developed a toe-probing strategy before stepping up. While this strategy is viable in simulation, this behavior results in hardware failures.

\subsection{Policy Training}
The Proximal Policy Optimization (PPO) algorithm \cite{schulman2017proximal} is used to learn the footstep controller in a Drake simulation environment \cite{drake}. To enhance the accuracy of the value estimation and speed up training, we employ an asymmetric actor-critic in which only the critic has access to the privileged information. Unlike the actor, the critic additionally receives the ground truth height map, joint positions, and pelvis pose. We include a symmetric mirror loss \cite{yu2018learning} to the original PPO objective function to discourage asymmetric footsteps and keep a healthy symmetric gait:
\begin{equation}
L_{mirror}(\theta) = w\sum_{i=0}^{N}
\| \pi_{\theta}(s_i) - \Psi_{act}\left(\pi_{\theta}\left(\Psi_{obs}(s_i)\right)\right) \|^2
\end{equation}
where $\pi_{\theta}$ is the policy, $s_i$ represents the $i$th observation, $\Psi_{act}(\cdot)$ and $\Psi_{obs}(\cdot)$ mirrors the actions and observations, and $N$ is the number of rollout samples. The coefficient $w$ defines the importance of symmetry, and $w=2$ is used for training.

We initialize the agent in a randomized pose and swing phase. The commanded velocities are sampled from $v_x \in [-0.8, 0.8]$, $v_y \in [-0.4, 0.4]$ for flat terrain and $v_x \in [0, 0.8]$, $v_y \in [-0.4, 0.4]$ for non-flat terrain. Backward walking is restricted to only flat terrains.

\begin{figure}[t]
    \centering
    \includegraphics[width=0.9\linewidth]{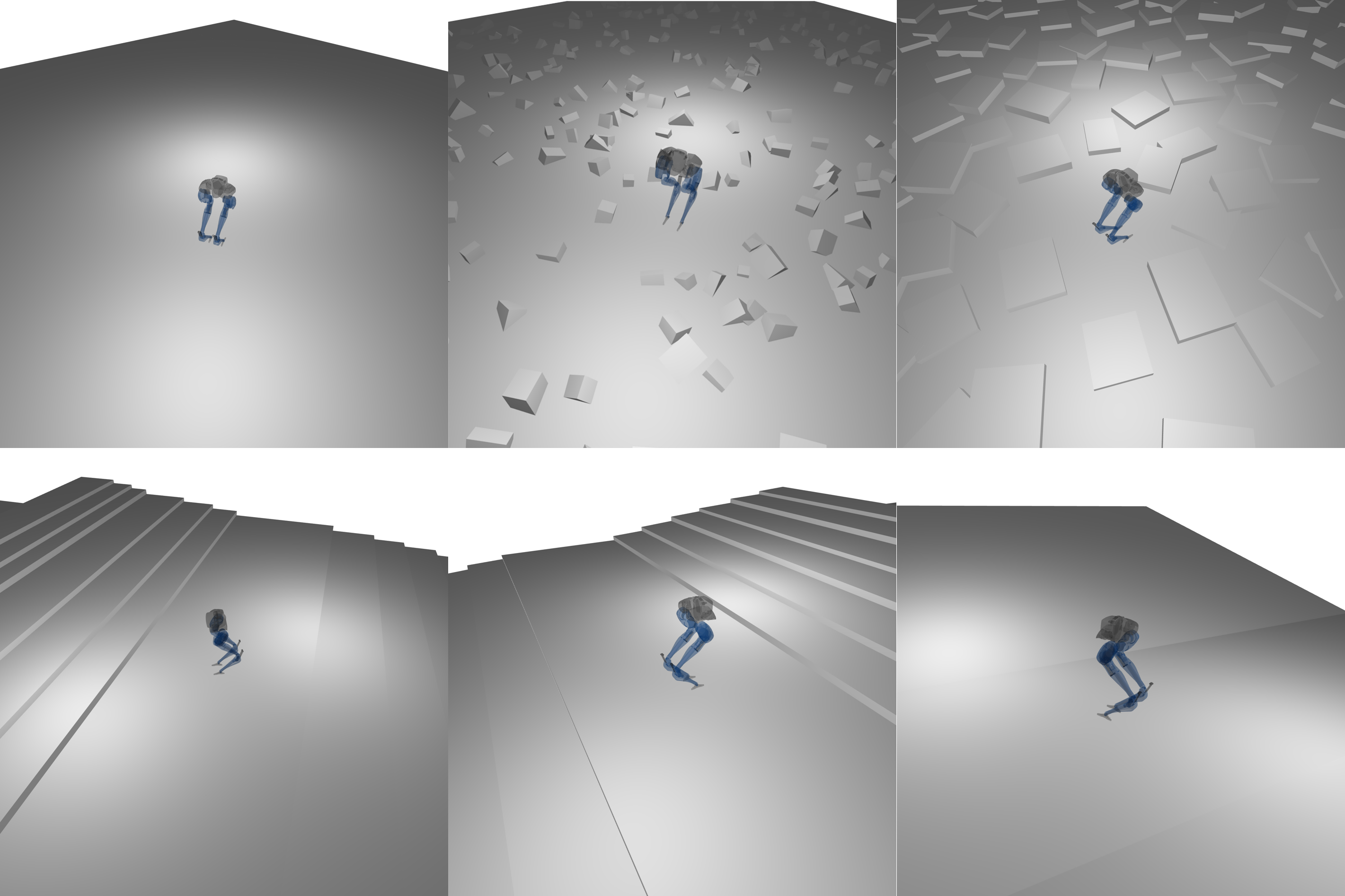}
    \caption{Terrain types used for training. Top row (left to right): flat, flat with obstacles, and block terrains. Bottom row (left to right): stairs, stairs with slopes, and sloped terrains.}
    \label{fig:terrain}
\end{figure}

The policy is trained across six terrain categories (Fig. \ref{fig:terrain}): (1) flat terrain; (2) flat terrain with randomly distributed obstacles of varying dimensions; (3) flat terrain with randomly distributed blocks of varying sizes; (4) stairs with varying width and height; (5) stairs with slopes; (6) slopes with varying angles.
Episodes are conducted for 400 timesteps, equivalent to 10 seconds in simulation time, across random terrains and modeling parameters. Early termination occurs under three conditions: (1) foot-to-pelvis distance is less than 20 cm, indicating a fall; (2) the magnitude of swing foot tracking error exceeds 50$\%$ or (3) self-collision.
\begin{table}[t]
\scriptsize
\centering
\caption{Terrain Categories}
% \vspace{-3mm}
\label{table:terrain}
\begin{tabular}{|c||c|c|}
\hline
\textbf{Terrain}  & \textbf{Parameters} & \textbf{Range}  \\ \hline
\textbf{Flat}              & $\times$            & $\times$        \\ \hline
\multirow{2}{*}{\textbf{Flat w/ Obstacle}}
& Obstacle Dimension XYZ [m] & $[0.2, 0.5]$\\ \cline{2-3}
& No. of Obstacle& $[30, 40]$\\ \hline
\multirow{3}{*}{\textbf{Block}}
& Block Dimension XY [m] & $[0.5, 1.0]$\\ \cline{2-3}
& Block Dimension Z [m] & $[0.05, 0.15]$\\ \cline{2-3}
& No. of Block & $[10, 20]$\\ \hline
\multirow{2}{*}{\textbf{Stair}}
& Stair Height [m] & $[0.075, 0.17]$\\ \cline{2-3}
& Stair Width [m] & $[0.5, 1.5]$\\ \hline
\multirow{3}{*}{\textbf{Stair w/ Slope}}
& Stair Height [m]& $[0.07, 0.14]$\\ \cline{2-3}
& Stair Width [m]& $[0.8, 1.6]$\\ \cline{2-3}
& Slope Angle [rad]& $ [0.03, 0.07]$\\ \hline
\textbf{Slope} & Slope Angle [rad] & [0.1, 0.34] \\ \hline
\end{tabular}
\vspace{-3mm}
\end{table}
\section{Sim-to-real Transfer}

\subsection{Domain Randomization}
To address modeling and measurement uncertainties between the simulated and real-world environments, we introduce randomization across dynamics parameters and observable states. Empirically, we found that randomizing all of the parameters in Table \ref{table:random} is necessary for sim-to-real. All randomization follows a uniform distribution sampled within a predefined range shown in Table \ref{table:random}. We introduce noise into the mass-normalized ALIP states to tackle discrepancies between simulation and hardware due to the differences in dynamics during ground contact. In addition, we perturb the elevation map through a combination of local and global displacements. At the beginning of each episode and at every timestep, the entire elevation map is shifted along the $x,y,z$ axes. Uniform noise is further applied to the elevation map to regularize the encoder. To emulate imperfections in real-world sensor calibration, we also introduce systematic biases in the point cloud data along the $x,y,z$ directions.

\begin{table}[t]
\scriptsize
\centering
\caption{Domain Randomization Parameters}
% \vspace{-3mm}
\label{table:random}
\begin{tabular}{|c||l|l|}
\hline
\textbf{} & \textbf{Parameters} & \textbf{Range}   \\ \hline
\multirow{5}{*}{\shortstack{\textbf{Dynamics} \\ \textbf{Model}}}
                         & PD Gains & $[0.5, 1.5] \times \text{Default}$\\
                         \cline{2-3} 
                         & Joint Damping & $[0.5, 2.5] \times \text{Default}$\\
                         \cline{2-3} 
                         & Link Mass & $[0.6, 1.4] \times \text{Default}$\\
                         \cline{2-3}
                         % & Joint position & $[-0.01, 0.01]\text{rad}$\\
                         \cline{2-3}
                         & ALIP state & $[-0.03, 0.03]$\\
                         \cline{2-3}
                         & Friction Coefficient & $[0.3, 1.1]$\\
                         \hline

\multirow{6}{*}{\shortstack{\textbf{Elevation} \\ \textbf{Map}}}
                         & Shift XY per episode      & $[-0.03, 0.03] \text{m}$ \\
                         \cline{2-3}
                         & Shift XY per timestep     & $[-0.02, 0.02]\text{m}$ \\
                         \cline{2-3}
                         & Shift Z per episode       & $[-0.02, 0.02] \text{m}$ \\
                         \cline{2-3}
                         & Shift Z per timestep      & $[-0.01, 0.01] \text{m}$ \\
                         \cline{2-3}
                         & Uniform Noise             & $[-0.02, 0.02]\text{m}$   \\ \cline{2-3}
                         & Point Cloud Bias XYZ             & $[-0.03, 0.03]\text{m}$ \\
                        \hline
\multirow{1}{*}{\textbf{Communication}}
                        & Delay         & $[0, 0.025]\text{s}$   \\
                        \hline
\multirow{2}{*}{\textbf{Perturbations}}
                        & Force XY on pelvis        & $[-20,20]\text{N}$ \\
                         \cline{2-3}
                        & Force Z on pelvis         & $[-10,10]\text{N}$   \\
                        \hline
\end{tabular}
\vspace{-5mm}
\end{table}

\subsection{Curriculum Learning}
We adopt a curriculum that adjusts the environment and domain randomization parameters for stable training. In the initial phase, training is conducted on flat terrain, staircases with a maximum height of 10 cm, and slopes. During this phase, domain randomization is excluded, as premature introduction of these elements causes policy divergence and cheating of the agent.
After convergence of the initial stage, we include all terrain types and apply domain randomization. We constrain the probability of the flat terrain at 10$\%$ to avoid catastrophic forgetting of learned negative velocity commands since the flat terrain is the only terrain type that associates with negative $x$ direction velocity.

\subsection{Elevation Map Drift Correction}
Due to the lack of direct global position sensing on the perception module, we observe consistent drift in the vertical ($z$) direction of the state estimator. This drift is primarily caused by impacts during foot contact, which perturb the floating base estimate over time, and as a result, the elevation map underestimates the terrain height.
To address this problem, a drift correction strategy based on the known position of the stance foot is used \cite{11077715}. Before each update of the map, we compute the vertical offset between the current stance foot position and the corresponding elevation in the map. This difference is applied as a correction offset and maintains alignment between the map and the ground surface.
\section{Simulation Experiments}

\begin{table*}[t]
\centering
\scriptsize
\caption{Performance comparison of RL policy architectures vs. MPC}
\label{tab:comparison}
% \vspace{-3mm}
\begin{tabular}{|c|c||c|c|c||c|c|c||c|c|c|}
\hline
\multirow{2}{*}{\textbf{Metric}} & \multirow{2}{*}{\textbf{Terrain Type}} & \multicolumn{3}{c|}{\textbf{ALIP RL}} & \multicolumn{3}{c|}{\textbf{Joint RL}} & \multicolumn{3}{c|}{\textbf{MPC \cite{11077715}}} \\ \cline{3-11} & & \textbf{$\mu$=0.4} & \textbf{$\mu$=0.8} & \textbf{$\mu$=1.1} & \textbf{$\mu$=0.4} & \textbf{$\mu$=0.8} & \textbf{$\mu$=1.1} & \textbf{$\mu$=0.4} & \textbf{$\mu$=0.8} & \textbf{$\mu$=1.1} \\ \hline \hline
\multirow{4}{*}{\shortstack{Mean Squared Error of \\ Velocity Tracking ($\text{m}^2/\text{s}^2)$}}
& Stair (ascend)   & 0.0388 & 0.0514 & 0.0505 & \textbf{0.0203} & \textbf{0.0209} & \textbf{0.0236} & 0.0834 & 0.0726 & 0.0725 \\ \cline{2-11}  
& Stair (descend)  & 0.0299 & 0.0351 & 0.0398 & \textbf{0.0228} & \textbf{0.0247} & \textbf{0.0260} & 0.0709 & 0.0665 & 0.0708 \\ \cline{2-11} 
& Slope (ascend)   & 0.0186 & 0.0153 & 0.0211 & \textbf{0.0046} & \textbf{0.0064} & \textbf{0.0072} & 0.0258 & 0.0213 & 0.0227 \\ \cline{2-11} 
& Slope (descend)  & 0.0424 & 0.0323 & 0.0346 & \textbf{0.0129} & \textbf{0.0104} & \textbf{0.0099} & 0.1166 & 0.0708 & 0.0654 \\ \hline \hline
\multirow{4}{*}{\shortstack{Success Rate (\%)}}
& Stair (ascend)   & 90.5 & 88 & 81.5 & \textbf{95.5} & \textbf{88.5} & \textbf{91.5} & 56 & 72.5 & 68.5 \\ \cline{2-11}  
& Stair (descend)  & \textbf{92} & \textbf{94.5} & \textbf{95} & 91.5 & 91.5 & 91 & 63.5 & 78 & 83 \\ \cline{2-11} 
& Slope (ascend)   & \textbf{100} & \textbf{100} & 96 & \textbf{100} & \textbf{100} & \textbf{100} & \textbf{100} & \textbf{100} & \textbf{100} \\ \cline{2-11} 
& Slope (descend)  & 82.5 & \textbf{100} & \textbf{100} & \textbf{99.5} & 99 & 99 & 5 & 80.5 & 96.5 \\ \hline
\end{tabular}
\vspace{-5mm}
\end{table*}

\subsection{Simulation Setup}
We trained two policies for analysis using identical network architecture, where the \emph{ALIP} policy refers to a model using an observation space that includes the elevation map, desired velocity commands, and the mass-normalized ALIP state (Fig. \ref{fig:overview}). The \emph{Joint} policy shares this observation space, with the addition of joint positions. Both policies are trained using all the methods discussed in this paper. In simulation, we compare \emph{ALIP}, \emph{Joint}, and a vision-based ALIP MPC footstep planner from \cite{11077715}; denoted as \emph{MPC}.

The controllers are evaluated using two primary metrics: \textit{velocity tracking accuracy} and \textit{success rates} across varying terrain types and friction coefficients. Stair terrains are generated using the parameters listed in Table \ref{table:terrain}, with a 17-degree incline used for the slope terrain and friction coefficients of 0.4, 0.8, and 1.1 as specified in Table \ref{tab:comparison}. For each experimental configuration, 200 episodes are collected with random target velocities. All evaluations are performed without any noise. An episode is considered successful if the robot does not fall for 15 seconds. 

\subsection{Simulation Evaluation}
\textit{1) Controller Comparison:} As shown in Table \ref{tab:comparison}, both the RL policies consistently outperform \emph{MPC} in velocity tracking across all terrain types and achieve higher success rates for ascending and descending stair and slope terrains. Furthermore, the \emph{Joint} policy demonstrates better performance in velocity tracking compared to the \emph{ALIP} policy across all of the evaluated terrains with similar success rates.

\begin{figure}[t]
    \centering
    \includegraphics[width=\linewidth]{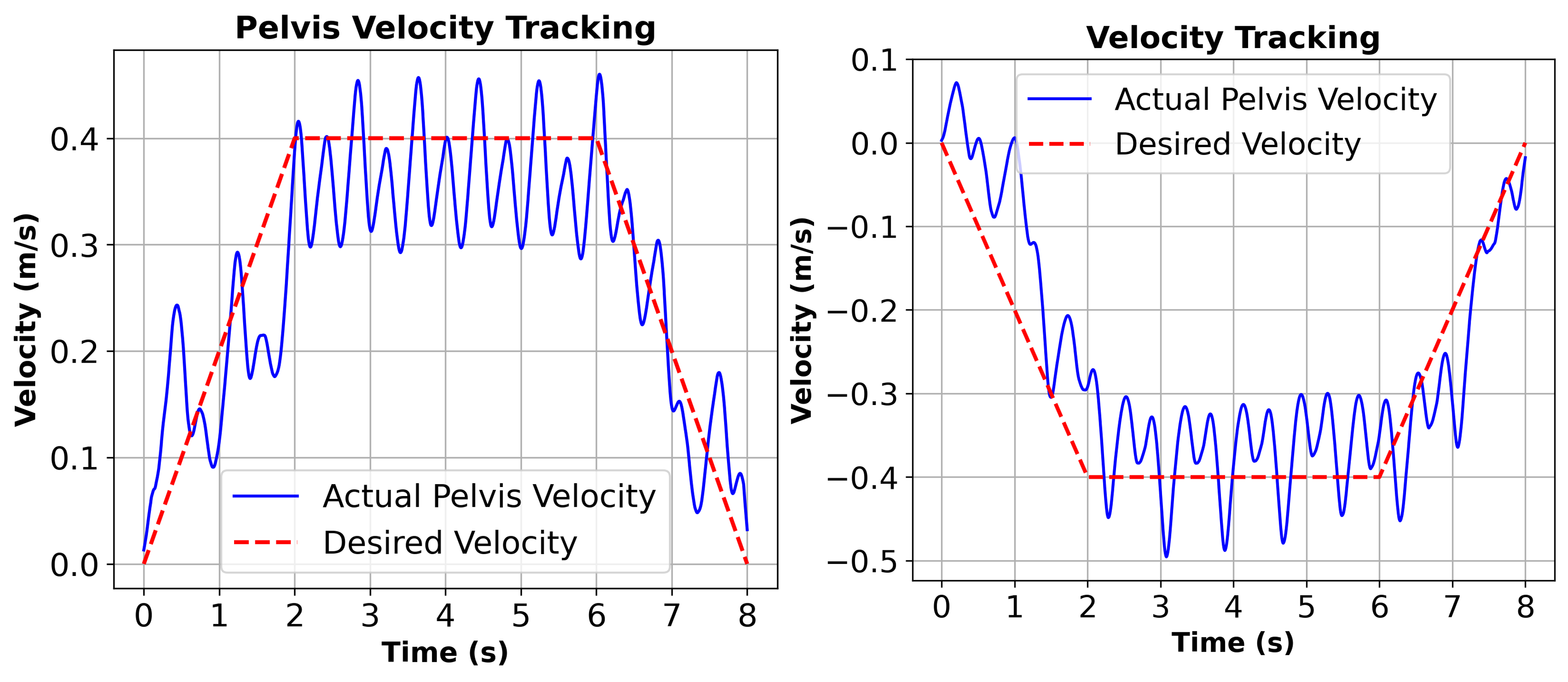}
    \vspace{-7mm}
    \caption{Velocity tracking performance in simulation using a predefined velocity profile for motion in the x direction on flat terrain.}
    \vspace{-5mm}
    \label{fig:velocity}
\end{figure}

\textit{2) Collision Evaluation:}
Hard collisions with staircases accounted for a large proportion of failures, contributing to the lower performance of \emph{MPC} on stair terrains. To further investigate, we assess the footstep planners based on footstep collisions with forces exceeding 1000 N. We quantify the frequency of such collisions and identify instances where these collisions directly lead to failure. A collision is classified as failing if the robot falls within two seconds of the impact.

Evaluation results indicate that the \emph{MPC} baseline exhibits a significantly higher frequency of hard collisions per episode compared to the RL-based planners. \emph{MPC} achieves an average of 0.4833 hard collisions per episode, while \emph{ALIP} and \emph{Joint} achieve lower rates of 0.1467 and 0.11, respectively.

\textit{3) Recovery:}
In addition to reducing the frequency of hard collisions, the RL-based policies demonstrate improved recovery capabilities following hard collision events. The failure rate following a collision for \emph{MPC} is 52.68$\%$, whereas \emph{ALIP} and \emph{Joint} show considerably lower failure rates of $38.88\%$ and $27.71\%$, respectively. Notably, recovery behaviors observed in simulation also transfer to hardware, where the RL-based policies show the ability to maintain balance and continue locomotion following disturbances.

\textit{4) Obstacle Avoidance:}
\begin{figure}[t]
    \centering
    \includegraphics[width=\linewidth]{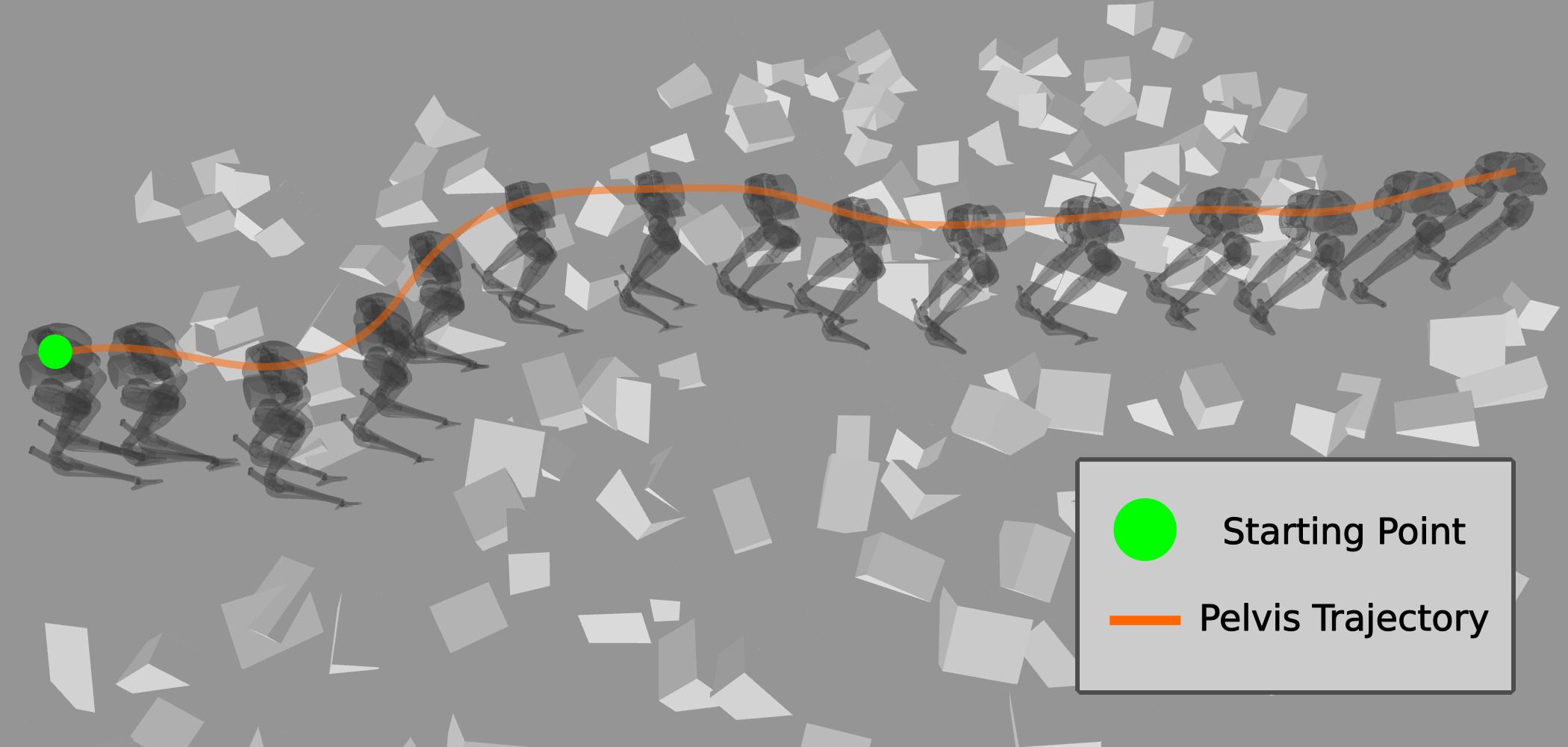}
    \vspace{-7mm}
    \caption{Given a forward velocity command of 0.4m/s and no lateral velocity input, the policy successfully avoids obstacles while moving forward.}
    \vspace{-5mm}
    \label{fig:obs}
\end{figure}
We demonstrate that incorporating obstacles directly into the simulation environment without introducing explicit safety or obstacle avoidance reward/penalty terms is sufficient for the policy to learn obstacle avoidance behaviors. Even in the absence of lateral velocity commands, the policy can navigate around obstacles while maintaining forward motion (Fig. \ref{fig:obs}). However, the policy may get stuck between obstacles, resulting in an in-place stepping behavior.

\section{Hardware Experiments}
\subsection{Hardware Setup}
The state estimator and operational space controller are executed on Cassie’s onboard NUC computer, and we transmit the joint torque commands to Cassie’s target PC via UDP. The RL footstep planner and perception module operate on a ThinkPad p15 Laptop, equipped with an 8-core, 2.3 GHz Intel 1180H processor and 24 GB of RAM. The laptop is carried by one of the safety bar carriers and is networked with the NUC over Ethernet for LCM \cite{huang2010lcm} communication.

\subsection{Hardware Evaluation}
\begin{figure*}[!t]
    \centering
    \includegraphics[width=\textwidth]{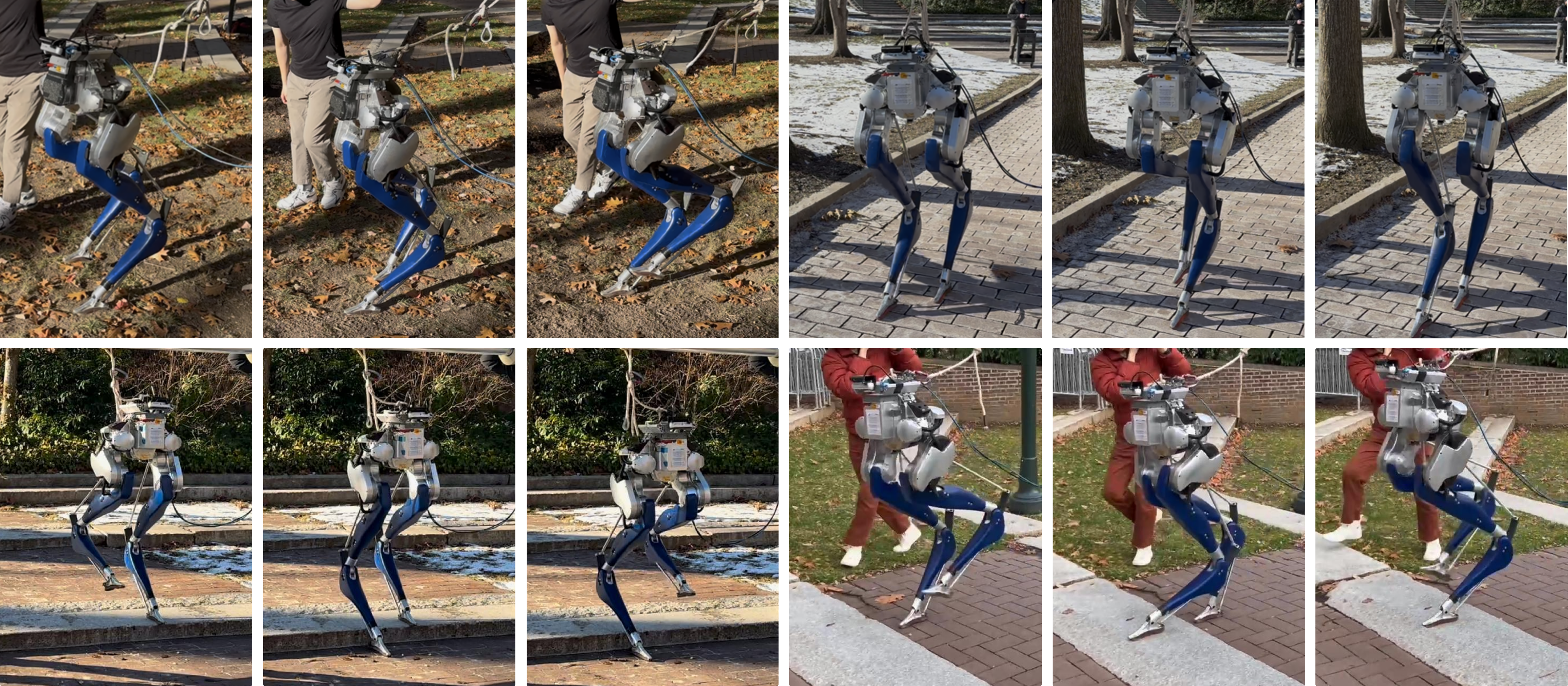}
    % \vspace{-7mm}
    \caption{Sequence of images of Cassie walking on outdoor terrains. Top Row: Walking on outdoor flat terrain. Bottom Row: Descending (left three) and ascending (right three) outdoor stair terrain.}
    \label{fig:motion_tile}
    \vspace{-3mm}
\end{figure*}

The policy transferred to hardware despite a bent right bar and a difference in pelvis mass, with the custom battery used for hardware testing being approximately 1 kg lighter than the nominal URDF model.
We demonstrate our walking controller across three scenarios: indoor flat terrain, outdoor flat-like terrain, and outdoor stairs with non-uniform, irregular slopes, all with perception in the loop. Trials are shown in Fig. \ref{fig:motion_tile}. The controller performs successfully on flat-like terrains in both indoor and outdoor environments. However, shows poor performance when ascending stairs, where \emph{ALIP} underperforms relative to \emph{Joint}, achieving a maximum of two steps before failure, compared to four steps with \emph{Joint}.

\begin{figure}[!t]
\centering
\includegraphics[width=\linewidth]{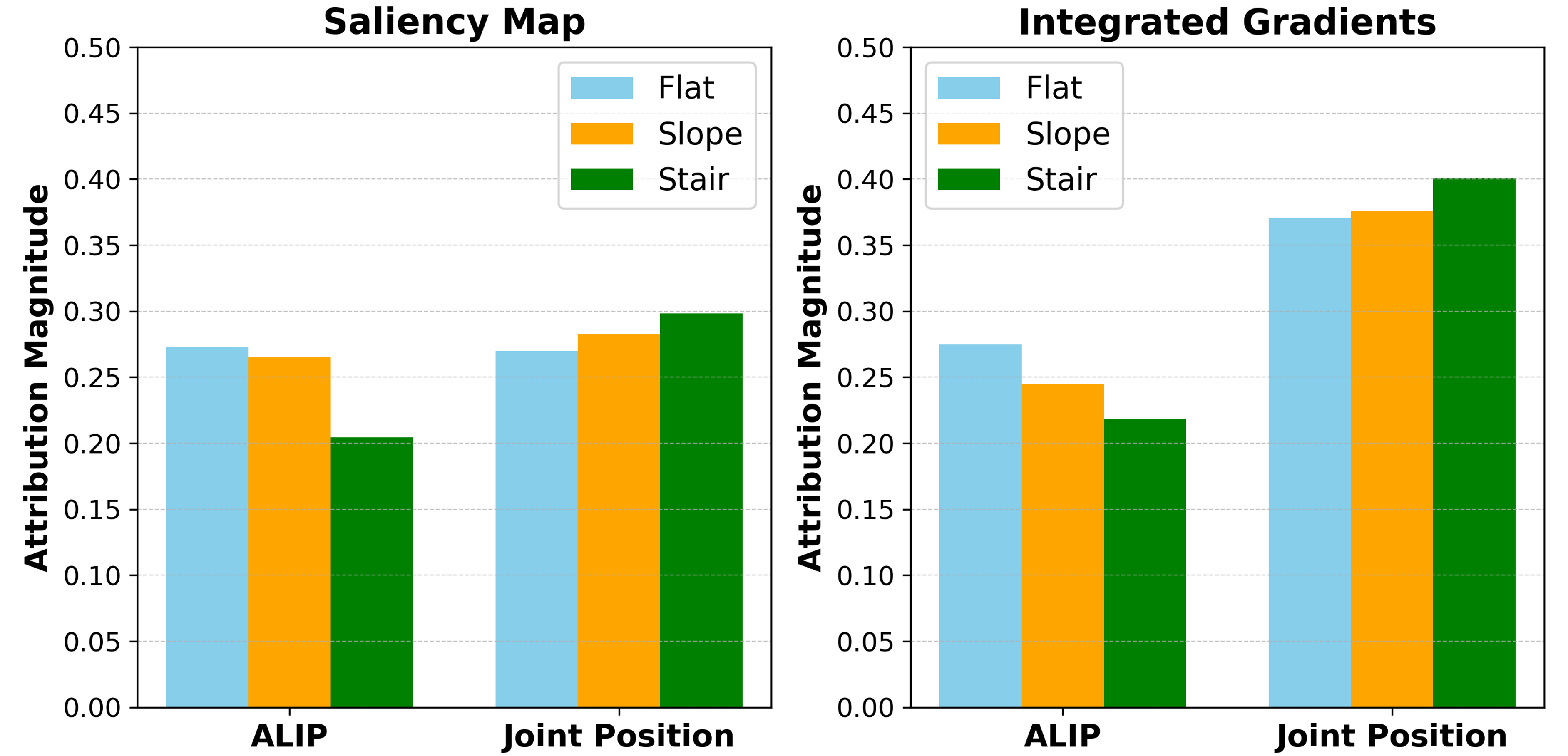}
\vspace{-7mm}
\caption{Feature attribution visualizations for flat, slope, and stair terrains on the \emph{Joint} policy. Left: Saliency maps, Right: Integrated gradients.}
\label{fig:attribution}
\vspace{-5mm}
\end{figure}

\section{Analysis}
\subsection{ALIP as a Policy Input}
We evaluate the comparative performance of the \emph{ALIP} and \emph{Joint} policies in stair-climbing tasks using the \textit{mean stairs-to-failure} metric. In this evaluation, episodes are extended to 50 seconds, and episodes are terminated if the robot falls, has hard collisions with the stairs, or deviates significantly in the lateral (y) direction. A total of 200 episodes were collected for each policy. \emph{Joint} averaged 10.024 steps before failure, outperforming \emph{ALIP}, which averaged 6.534 steps. This is consistent with hardware observations and reflects the limitations of the ALIP model in handling height variations.

To assess the importance of ALIP states in policy decision-making, we employ two attribution methods: saliency maps \cite{simonyan2013deep} and integrated gradients \cite{sundararajan2017axiomatic}.
As shown in Fig.~\ref{fig:attribution}, the influence of joint position inputs increases, while the contribution of ALIP state decreases with increasing terrain complexity in the \emph{Joint} policy. This suggests the need for more expressive inputs in challenging discontinuous environments such as stair climbing. Our observation aligns with the findings of \cite{castillo2023template}, which demonstrate successful policy performance on sloped terrains but do not report evaluations on discontinuous or stair-like environments, which are likely to be less successful due to the limitations of the ALIP reduced-order model.

\begin{table}[t]
\centering
\caption{L2 ratio of OSC output tracking $\downarrow$}
\vspace{-3mm}
\label{tab:l2_ratios}
\resizebox{\linewidth}{!}{%
\begin{tabular}{|c||c|c|c||c|c|c|}
\hline
\multirow{2}{*}{\shortstack{Trajectory}}
& \multirow{2}{*}{\shortstack{RL \\ Real}}
& \multirow{2}{*}{\shortstack{RL \\ Sim}}
& \multirow{2}{*}{\shortstack{Sim2real \\ Error (RL)}}
& \multirow{2}{*}{\shortstack{MPC \\ Real}}
& \multirow{2}{*}{\shortstack{MPC \\ Sim}}
& \multirow{2}{*}{\shortstack{Sim2real \\ Error (MPC) }} \\
 &  &  &  &  &  & \\ \hline
Swing Foot $x$       & 0.2448 & 0.1698 & 0.075 & 0.1960 & 0.1279 & 0.0681 \\
Swing Foot $\dot{x}$ & 0.7482 & 0.4966 & 0.2516 & 0.4921 & 0.4874 & 0.0047 \\
Swing Foot $y$       & 0.0688 & 0.0529 & 0.0159 & 0.0607 & 0.0415 & 0.0192 \\
Swing Foot $\dot{y}$ & 1.3454 & 0.1747 & 1.1707 & 1.0785 & 0.2417 & 0.8368\\
Swing Foot $z$       & 0.1440 & 0.1274 & 0.0166 & 0.1117 & 0.1169 & 0.0052 \\
Swing Foot $\dot{z}$ & 0.3265 & 0.3084 & 0.0181 & 0.3036 & 0.2288 & 0.0748 \\
\hline
CoM $z$       & 0.0179 & 0.0119 & 0.0060 & 0.0084 & 0.0094 & 0.0010\\
CoM $\dot{z}$ & 3.9889 & 3.0995 & 0.8894 & 2.2148 & 3.0187 & 0.8039 \\
\hline
\end{tabular}}
\vspace{-5mm}
\end{table}

\subsection{Hierarchical Control}
We evaluate the tracking performance of the operational space controller (OSC) using reference trajectories from the \emph{MPC} and \emph{Joint} footstep planners, averaged over four stair ascent trials in both simulation and hardware (Table.~\ref{tab:l2_ratios}). To quantify performance, we report the L2 ratio, defined as:
\begin{equation}
\text{L2 Ratio} = \frac{\sqrt{\sum_t \|x_{\text{err}}(t)\|^2}}{\sqrt{\sum_t \|x_{\text{ref}}(t)\|^2}}
\end{equation}
where $x_{err}(t)$ is the tracking error and $x_{ref}(t)$ is the reference trajectory at time $t$. The scale-invariant metric allows comparison across trials with different trajectory magnitudes. Ideally, if the OSC successfully feedback linearized the output, the L2 ratio would remain invariant to the reference signal. However, Table~\ref{tab:l2_ratios} shows this is not the case. While both planners yield relatively low and comparable L2 ratios in simulation, the RL planner consistently results in higher L2 ratios across all output components in hardware.

This indicates that in the real-world, while the MPC planner produces trajectories that remain within the effective operating range of the OSC, the RL policy outputs footstep positions that drive the OSC into domains where the performance degrades. This effect is seen in hardware, where the sim-to-real gap further exacerbates the degradation in OSC performance, as reflected by the elevated L2 ratios.

\section{Conclusion and Future Works}
We present a learned hierarchical control framework for vision-based bipedal locomotion. The RL footstep planner utilizes a reduced-order ALIP model and an elevation map derived from a single depth camera. By simplifying both the observation and action spaces, we reduce the overall problem complexity. We validated the framework through simulation and hardware experiments, and interpreted the collected data to identify limitations of reduced-order models and hierarchical control within the context of hierarchical reinforcement learning frameworks.

While the use of the ALIP model and a hierarchical architecture offers benefits in terms of model simplicity and training efficiency, we show their limitations in handling complex tasks and real-world deployment. The ALIP-based footstep planner underperforms in complex environments due to its limited representational capacity. Additionally, although hierarchical controllers are known to improve training efficiency, interpretability, and modularity \cite{bao2024deep}, our results indicate they complicate sim-to-real transfer. Transferring policies across layers increases the overall complexity of the transfer process, and the hierarchical structure imposes communication dependencies between layers, which require precise modeling and can hinder policy generalization.

Future work will explore more expressive reduced-order models for handling discontinuous terrain and focus on improving the alignment between the learned high-level policy and the low-level controller to enhance sim-to-real transfer.

\bibliographystyle{IEEEtran}
\bibliography{bib}
\end{document}